\documentclass[a4paper,twoside]{article}

\usepackage{epsfig}
\usepackage{booktabs}
\usepackage{graphicx}

\usepackage{calc}
\usepackage{amssymb}
\usepackage{amstext}
\usepackage{amsmath}
\usepackage{amsthm}
\usepackage{multicol}
\usepackage{pslatex}
\usepackage{apalike}
\usepackage{verbatim}

\usepackage{SCITEPRESS}     

\begin{document}

\bibliographystyle{apalike}

\title{ A Comparison of Embedded Deep Learning Methods for Person Detection }

\author{\authorname{Chloe Eunhyang Kim\sup{1}, Mahdi Maktab Dar Oghaz\sup{2}, Jiri Fajtl\sup{2}, Vasileios Argyriou\sup{2}, Paolo Remagnino\sup{2}}
\affiliation{\sup{1}VCA Technology Ltd, Surrey, United Kingdom}
\affiliation{\sup{2}Kingston University, London, United Kingdom}
\email{chloe.kim@vcatechnology.com\{m.maktabdaroghaz , j.fajtl , vasileios.argyriou , 
p.remagnino\}@kingston.ac.uk}
}


\abstract{Recent advancements in parallel computing, GPU technology and deep learning provide a new platform for complex image processing tasks such as person detection to flourish. Person detection is fundamental preliminary operation for several high level computer vision tasks.
One industry that can significantly benefit from person detection is retail. 
In recent years, various studies attempt to find an optimal solution for person detection using neural networks and deep learning.
This study conducts a comparison among the state of the art deep learning base object detector with the focus on person detection performance in indoor environments.
Performance of various implementations of YOLO, SSD, RCNN, R-FCN and SqueezeDet have been assessed using our in-house proprietary dataset which consists of over 10 thousands indoor images captured form shopping malls, retails and stores.
Experimental results indicate that, Tiny YOLO-416 and SSD (VGG-300) are the fastest and Faster-RCNN (Inception ResNet-v2) and R-FCN (ResNet-101) are the most accurate detectors investigated in this study.
Further analysis shows that YOLO v3-416 delivers relatively accurate result in a reasonable amount of time, which makes it an ideal model for person detection in embedded platforms.}

\onecolumn \maketitle \normalsize \vfill

\section{\uppercase{Introduction}}
\label{sec:introduction}

\noindent The rise of industry 4.0, IoT and embedded systems pushes various industries toward data driven solutions to stay relevant and competitive. In the retail industry, customer behavior analytic is one of the key elements of data driven marketing. Metrics such as customer's age, gender, shopping habits and moving patterns allow retailers to understand who their customers are, what they do and what they are looking for. these metrics also enables retailers to push customized and personalized marketing schemes to their customers across various stages of the customer life-cycle. Additionally, with the help of predictive models, retailers are now enable to predict what their customers are likely to do in the future and gain edge over their competitors. In recent years, there has been an increasing interest in the analysis of in-store customer behavior. Retailers are looking for insights on in-store customer's journey; Where do they go? What products do they browse? and most importantly, which products do they purchase \cite{ghosh2017acquiring} \cite{majeed2017internet} \cite{balaji2017value}? 

Over the last decade, several tracking approaches such as sensor based, optical based and radio based have been proposed. However, The majority of them are not efficient and reliable enough, or they expect some form of interaction with customers which might compromise their shopping experience \cite{jia2016cooperative}\cite{foxlin2014motion}. Analysis of in-store customer behavior through optical video signal recorded by security cameras has clear advantage over other approaches as it utilizes the existing surveillance infrastructure and operates seamlessly with no interaction and interference with customers \cite{ohata2014analysis}\cite{zuo2016prediction}. Despite the clear advantage of this approach, analysis of video signal requires complex and computationally expensive models, which up until recent years, was impractical in the real world. Recent advancements in parallel computing and GPU technology diminished this computational barrier and allowed complex models such as deep learning to flourish \cite{nickolls2010gpu}.

Aside from hardware limitations, classic computer vision and machine learning techniques had hard time to model these complex patterns, however the rise of data driven approaches such as deep learning, simplified these tasks, eliminating the need for domain expertise and hard-core feature extraction.
A reliable yet computationally reasonable person detection model is fundamental requirement for in-store customer behavior analysis. Numerous studies focused on person detection using deep neural network models. However, none of which particularly focused on the person detection in in door retail environments. Despite the similarity of these topics, there are a number of unique challenges, such as lighting condition, camera angles, clutter and queues in retail environments, which questions the adaptability of the existing person detection solutions for retail environments.

In this regard, this research is mainly focused on person detection as a preliminary step for in-store customer behavior modeling. We are particularly interested in evaluation and comparison of deep neural network (DNN) person detection models in cost-effective, end-to-end embedded platforms such as the Jetson TX2 and Movidius. State of the art deep learning models use general purpose datasets such as PASCAL VOC or MS COCO to train and evaluate. Despite their similarities, these dataset cannot be true representative of the retail and store environments. In data driven techniques such as deep learning, this adaptability issues are more pronounced than ever before \cite{lecun2015deep}. To address these issues, this research investigates the performance of state of the art DNN models including variations of YOLO, SSD, RCNN, R-FCN and SqueezeDet in person detection using an in-house proprietary image dataset were captured by conventional security cameras in retail and stores environments.

These images were manually annotated to form the ground truth for training and evaluation of the deep models. Having deep models trained by the similar type of images that could be found in target environment, can significantly improve the accuracy of the models. However, preparation of a very large annotated dataset is a big challenge. This research employs \textit{average precision} metric at various \textit{intersection over union} (IoU) as the figure of merit to compare model performance. As processing speed is a key factor in embedded systems, this research also conducts a comprehensive comparison among the aforementioned DNN techniques to find the most cost-effective approach for person detection in embedded systems.

The major contributions of this study can be summarized as: first, integration and optimization of the state of the art person detection algorithm into embedded platforms; second, an end-to-end comparative study among the existing person detection models in terms of accuracy and performance and finally, a proprietary dataset, which can be used in indoor human  and analysis studies.

The paper is organized as follow.  Section 2 briefly describes the state of art object detection models  used in this research. Section 3 presents the overall framework, data acquisition process as well as experimental setup of the research. Section 4 describes the experimental results and discussions and finally, sections 5 concludes the research.

\section{\uppercase{CNN based Object Detection}}

\noindent Various DNN based object detector have been proposed in the last few years. This research investigates the performance of state of the art DNN models including variations of YOLO, SSD, RCNN, R-FCN and SqueezeDet in person detection. The models have been trained using an in-house proprietary image dataset were captured by conventional security cameras in retail and stores environments. The following sections describes aforementioned DNN models in more details.

\subsection{RCNN Variants}

\noindent The region-based convolutional neural network (RCNN) solution for object detection is quite straightforward. This technique uses selective search to extract just 2000 regions (region proposal) from the image and then, instead of trying to classify a huge number of regions throughout the image only these 2000 region will be investigated. Selective search initially generates candidate regions, then uses a greedy algorithm to recursively combine similar regions into larger ones. Finally, it uses the generated regions to produce the final candidate region proposals. The region proposals will be passed to a conventional neural network (CNN) for classification. Despite RCNN has lots of advantages over the conventional DNN object detector \cite{girshick2016region}, this technique is still quite slow for any real-time application. Furthermore, a predefined threshold of 2000 region proposal cannot be suitable for any given input image.

To address these limitations, other variants of RCNN have been introduced \cite{ren2015faster}. Faster RCNN is one popular variant of RCNN which mainly devised to speed up RCNN. This algorithm eliminates the selective search algorithm used in the conventional RCNN and allows the network learn the region proposals. The mechanism is very similar to fast RCNN where an image is provided as input to a CNN to generate a feature map but, instead of using a selective search algorithm on the feature map to identify the region proposals, a separate network is used to predict region proposals. The predicted region proposals are then reshaped using a region of interest (RoI) pooling layer and used to classify the image input within the proposed region \cite{ren2015faster}. To train the Region Proposal Network, a binary class label has been assigned to each anchor (1: being object and 0: not object). Any with IoU over 0.7 determines object presence and anything below 0.3 indicates no object exists. With these assumptions, we minimize an objective function following the multi-task loss in Fast R-CNN which is defined as:

\begin{align}
L(\{p_i\},\{t_i\})=\frac{1}{N_{cls}}\sum_{i}{N_{cls}}(p_i,{p_i}^*)+\notag\\
 \lambda\frac{1}{N_{reg}}\sum_{i}{{P_i}^*}L_{reg}(t_i,{t_i}*)
\end{align}

\noindent where $i$ is the index of anchor in the batch, $p_i$ is its predicted probability of being an object; ${p_i}^*$ is the ground truth probability of the anchor (1: represents object, 0: represents non-object); $t_i$ is a vector which denotes the bounding box coordinates; ${t_i}^*$ is ground truth bounding box coordinates; $L_{cls}$ is classification log loss and $L{reg}$ is regression loss. We have also deployed the Faster RCNN model using the Google inception framework which is expected to be less computational intensive.

\subsection{R-FCN Variants}

\noindent In contrast to the RCNN model which applies a costly per-region subnetwork hundreds of times, region based fully convolutional network (R-FCN) is an accurate and efficient object detector that spreads the computation across the entire image. A position-sensitive score map is used to find a tradeoff between translation-invariance in image classification and translation-variance in object detection. A position-sensitive score defined as following:

\begin{align}
r(i,j\mid\theta)=\sum_{{(x,y)}\in bin(i,j)}{z_{i,j,c}}(x+x_0 , y+y_0\mid\theta)/n
\end{align}

\noindent where $r_c(i,j)$ is the pooled response in the $(i,j)^{th}$ bin in the $c^{th}$ category; $z_{i,j,c}$ is one score map out of the $k^2(C+1)$ score map; $n$ in the number of the pixels in the bin; $(x_0,u_0)$ represents the top left corner of the region of interest and $\theta$ denotes network learning parameters. The loss function defined on each region of interest which calculated by summation of the cross entropy loss and box regression loss as following: 

\begin{align}
L(s,t_{x,y,w,h})=L_{cls}(s_{c^*})+ \lambda[c^*>0]L_{reg}(t,t^*)
\end{align}

\noindent where $c^*$ is the region of interest ground truth label; $L_{cls}(s_{c^*})$ is cross entropy loss for classification; $t^*$ represents the ground truth box and $L{reg}$ is the bounding box regression loss. Aside from the original R-FCN, this study also investigates the R-FCN model with the Google inception framework \cite{dai2016r}. 

\subsection{YOLO Variants}

\noindent You only look once (YOLO) is another state of the art object detection algorithm which mainly targets real time applications. it looks at the whole image at test time and its predictions are informed by global context in the image. It also makes predictions with a single network evaluation unlike models such RCNN, which require thousands for a single image. YOLO divides the input image into an $SxS$ grid. If the center of an object falls into a grid cell, that cell is responsible for detecting that object. Each grid cell predicts five bounding boxes as well as confidence score for those boxes. The score reflects how confident the model is about the presence of an object in the box. For each bounding box, the cell also predicts a class. It gives a probability distribution score over all the possible classes designate the object class. Combination of the confidence score for the bounding box and the class prediction, indicates the probability that this bounding box contains a specific type of object. The loss function is defined as:

\begin{align}
\lambda_{coord}\sum_{i=0}^{s^2}\sum_{j=0}^{B}1_{ij}^{obj}[(x_i-\widehat{x_i})^2+(y_i-\widehat{y_i})^2]+\notag\\
\lambda_{coord}\sum_{i=0}^{s^2}\sum_{j=0}^{B}1_{ij}^{obj}[(\sqrt{w_i}-\sqrt{\widehat{w_i}})^2+(\sqrt{h_i}-\sqrt{\widehat{h_i}})^2]+\notag\\
\sum_{i=0}^{s^2}\sum_{j=0}^{B}1_{ij}^{obj}(C_i-\widehat{C_i})^2+\lambda_{coord}\sum_{i=0}^{s^2}\sum_{j=0}^{B}1_{ij}^{obj}(C_i-\widehat{C_i})^2+\notag\\
\sum_{i=0}^{s^2}1_{ij}^{obj}\sum_{c\in{classes}}(p_i(c)-\widehat{p_i}(c))^2
\end{align}

\noindent where $1_{i}^{obj}$ indicates if object appears in cell $i$ and $1_{ij}^{obj}$ denotes the $j^{th}$ bounding box predictor in cell $i$ responsible for that prediction; $x,y,w,h$ and $C$ denote the coordinates represent the center of the box relative to the bounds of the grid cell, the width and height are predicted relative to the whole image and finally $C$ denotes the confidence prediction represents the IoU between the predicted box and any ground truth box. This study also investigates the other variants of YOLO including YOLO-v2 as well as Tiny YOLO models performance for person detection in retail environments \cite{redmon2016you}\cite{redmon2017yolo9000}.

\subsection{SSD Variants}

\noindent Single shot multi-box detector (SSD) is one of the best object detector in terms of speed and accuracy. The SSD object detector comprises two main steps including \textit{feature maps extraction}, and \textit{convolution filters application} to detect objects. A predefined bounding box (prior) is matched to the ground truth objects based on IoU ratio. Each element of the feature map has a number of default boxes associated with it. Any default box with an IoU of 0.5 or greater with a ground truth box is considered a match. For each box, the SSD network computes two critical components including
\textit{confidence loss} which measures how confident the network is at the presence of an object in the computed bounding box using categorical cross-entropy and \textit{location loss} which computes how far away the networks predicted bounding boxes are from the ground truth ones based on the training data \cite{huang2017speed}\cite{liu2016ssd}. The overall loss function is defined as following:
\begin{align}
L(x,c,l,g)=\frac{1}{N}(L_{conf}(x,c)+\alpha L_{loc}(x,l,g))
\end{align}
\noindent where $N$ is the number of matched default boxes. Other variants of the standard SSD with 300 and 512 inputs as well as MobileNet and Inception models has been implemented and tested in this research \cite{howard2017mobilenets}\cite{szegedy2015going}.
\subsection{SqueezeDet}
\noindent SqueezeDet is a real-time object detector used for autonomous driving systems. This model claims high accuracy as well as reasonable response latency, which are crucial for autonomous driving systems. Inspired by the YOLO, this model uses fully convolutional layers not only to extract feature maps, but also to compute the bounding boxes and predict the object classes. The detection pipeline of SqueezeDet only contains a single forward pass over the network, making it extremely fast \cite{wu2017squeezedet}. SqueezeDet can be trained end-to-end, similarly to the YOLO and it shares similar loss function with YOLO object detection. 

\section{\uppercase{Research Framework}}
\noindent Similar to any other machine learning task, this research employs training/testing and validation strategy to create the prediction models. All CNN models were trained and tested using our proprietary dataset. Predictions were compared against ground truth by means of cross entropy loss function to back propagate and optimize network weights, biases and other network parameters. Finally, the trained models were tested against an unseen validation set to identify the models performance in real life. Figure 1 shows overall experimental framework.

\begin{figure}[!h]
  \centering
   {\epsfig{file = 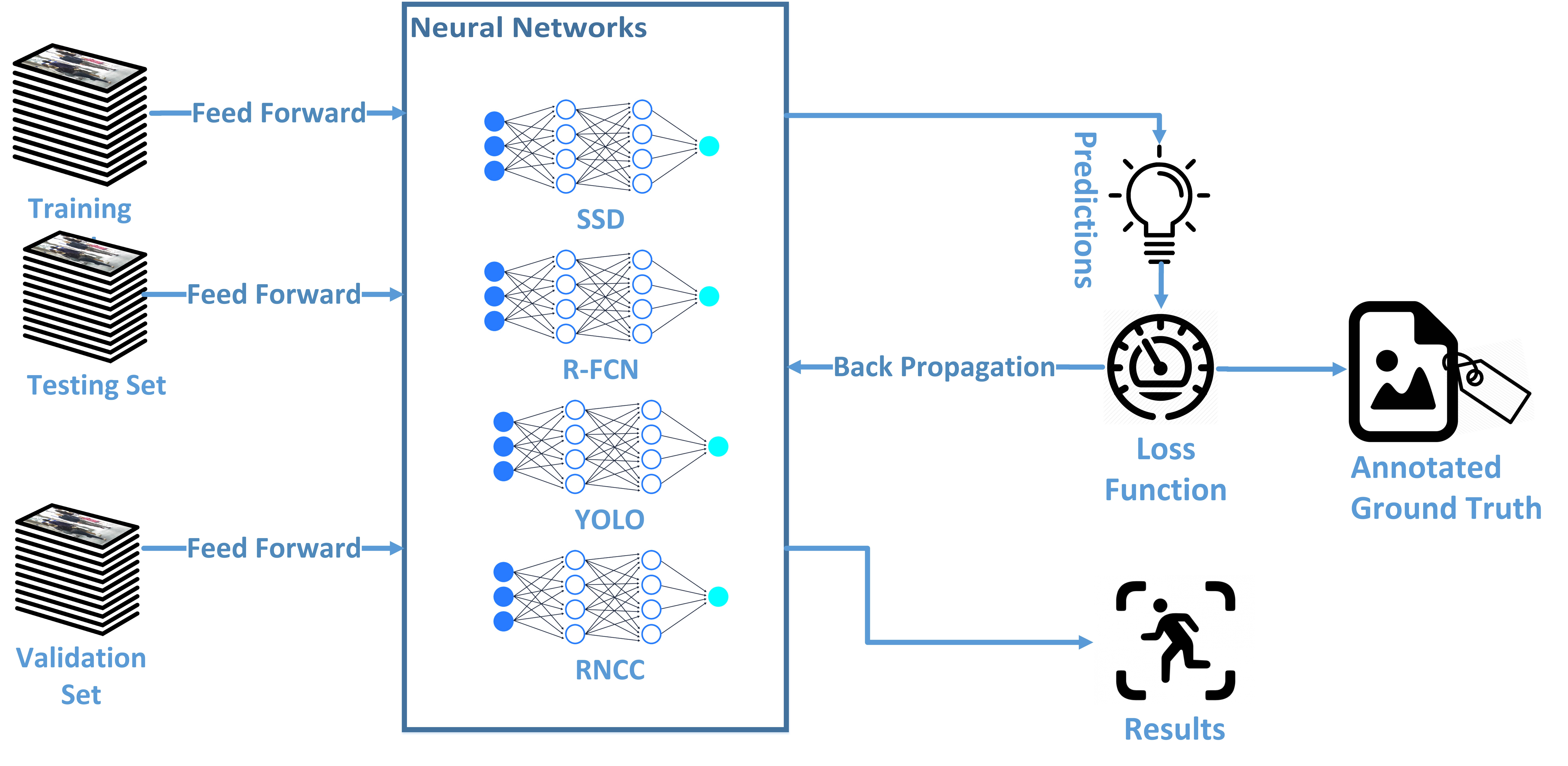, width = 7.5cm}}
  \caption{Overall experimental framework}
  \label{fig:example1}
 \end{figure}

\subsection{Data Acquisition}
\noindent 

We have prepared a relatively large dataset comprising total number of 10,972 image were mostly captured from CCTV cameras placed in department stores, shopping malls and retails. Majority of the images were captured in indoor environments under various conditions such as distance, lighting, angle, and camera type. Given the fact that each camera has its own color depth and temperature, field of view and resolution, all images passed through a preprocessing operation which ensures consistency across entire input data. Figure 2 shows some examples of our dataset.

\begin{figure}[!h]
  \centering
   {\epsfig{file = 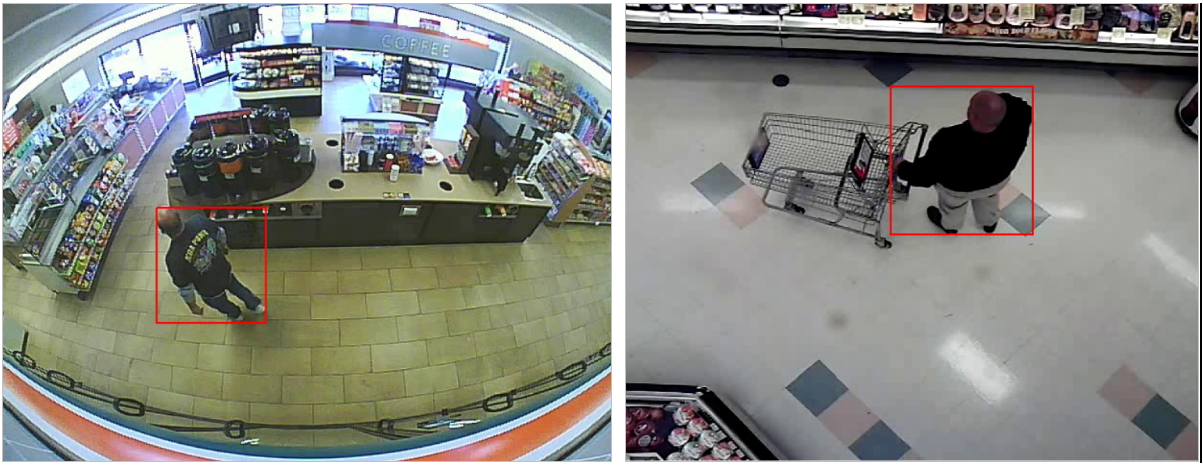, width = 7.5cm}}
  \caption{An example of tilt (left) and top-down (right) frame in dataset}
  \label{fig:example1}
 \end{figure}

In order to ease and speed up the annotation process, we have employed a semi-automatic annotation mechanism which uses a Faster RCNN inception model to generate the initial annotations for each given input image. The detection results were manually investigated and fine tuned to insure the reliability and integrity of the ground truth. Moreover, images with no person presence have been removed from the dataset. Finally, a random sampling process performed over entire images. The final dataset consists of total number of 10,972 images no background overlap, divided into training set (5,790 images), testing set (2,152 images) and validation set (3,030 images).

\subsection{Experimental Setup}
\noindent To measure and compare the average precision (AP) and IoU of the deep models, we have used a workstation powered by 16 GB of internal memory and Nvidia GTX 1080ti graphics accelerator. To measure and compare the time complexity metrics, we have utilized two common embedded platforms including the Nvidia Jetson TX2 as well as Movidius to run the experiments.  
\section{\uppercase{Experimental Results and Discussions}}
\noindent We investigated 13 different object detector deep models including variants of YOLO, SSD, RCNN, RFCN and SqueezeDet. To measure the accuracy of these
models, we have used AP at two different IoU ratios, including 0.5 which denotes a fair detection and 0.95 which indicates a very accurate detection. Table 2 summarizes the AP across various object detectors. It can be observed that, when IoU is 0.95, Faster RCNN (Inception ResNet-v2) with average precision of 0.317 outperforms other object detector in this research. Faster RCNN (ResNet-101) alongside R-FCN (ResNet-101) with respective AP of 0.245 and 0.246 are among the best performers in this category.

\begin{table}[]
\centering
\caption{Average precision at IoU 0.95 and 0.50}
\label{my-label}
\resizebox{\textwidth/2}{!}{%
\begin{tabular}{@{}lllll@{}}
\toprule
\# & Model & \multicolumn{1}{c}{Framework} & \multicolumn{1}{c}{\begin{tabular}[c]{@{}c@{}}AP\\  {[}IoU=0.95{]}\end{tabular}} & \multicolumn{1}{c}{\begin{tabular}[c]{@{}c@{}}AP\\  {[}IoU=0.50{]}\end{tabular}} \\ \midrule
1 & Faster RCNN (ResNet-101) & Tensorflow & \textbf{0.245} & \textbf{0.476} \\
2 & YOLOv3-416 & Darknet & 0.143 & 0.367 \\
3 & Faster RCNN (Inception ResNet-v2) & Tensorflow & \textbf{0.317} & \textbf{0.557} \\
4 & YOLOv2-608 & Darknet & 0.198 & \textbf{0.463} \\
5 & Tiny YOLO-416 & Darknet & 0.035 & 0.116 \\
6 & SSD (Mobilenet v1) & Tensorflow & 0.094 & 0.233 \\
7 & SSD (VGG-300) & Tensorflow & 0.148 & 0.307 \\
8 & SSD (VGG-500) & Tensorflow & 0.183 & 0.403 \\
9 & R-FCN (ResNet-101) & Tensorflow & \textbf{0.246} & \textbf{0.486} \\
10 & Tiny YOLO-608 & Darknet & 0.06 & 0.185 \\
11 & SSD (Inception ResNet-v2) & Tensorflow & 0.116 & 0.267 \\
12 & SqueezeDet & Tensorflow & 0.003 & 0.012 \\
13 & R-FCN & Tensorflow & 0.124 & 0.319 \\ \bottomrule
\end{tabular}%
}
\end{table}

On the other hand, SqueezeDet and Tiny YOLO-608 with respective AP of 0.003 and 0.06 performed poorly in this category. Results with IoU = 0.50 show a very similar trend. Once again, Faster RCNN (Inception ResNet-v2) with AP 0.557 outperformed other detector. R-FCN (ResNet-101), Faster RCNN (ResNet-101) and YOLOv2-608 with average precision of 0.486, 0.476 and 0.463 respectively, are showing superior performance. In contrast, SqueezeDet and Tiny YOLO-416 with respective AP of 0.012 and 0.116 generate poor results. Results also indicates, that, in terms of robustness and resiliency of the detector against increase in IoU, all models perform roughly equally and there is no significant variance. Another noteworthy observation in this experiment is the superiority of the Faster RCNN over other detectors that could be influenced biased by the approach used to prepared the ground truth. As we mentioned earlier in section 3.1, the dataset annotation initialized with the help of Faster RCNN inception model detector. Despite the significant manual adjustments and fine-tuning in annotation, we believe it introduces some level of bias to the results. 

The time complexity of detectors were evaluated with measurement of execution latencies’ in two different approaches. In the first approach total latency of inference of a single test image has been measured in both CPU and GPU modes. In the second approach throughput of continuous inference with repeating camera capture. Table 3 shows the total latency of inference of a single test image on both CPU and GPU. Apparently, GPU is considerably faster than a CPU in matrix arithmetics such as convolution due to their high bandwidth and parallel computing capabilities, but it is always interesting to learn this advantage objectively. According to the results shown in table 3, in CPU mode, SqueezeDet, SSD (Inception ResNet-v2) and SSD (Mobilenet-v1) are the fastest deep models in this study.

\begin{table}[]
\centering
\caption{Total latency of inference in both CPU and GPU modes}
\label{my-label}
\resizebox{\textwidth/2}{!}{%
\begin{tabular}{@{}llll@{}}
\toprule
\# & Model & CPU Latency (S) & GPULatency (S) \\ \midrule
1 & Faster RCNN (ResNet-101) & 3.271 & 0.232 \\
2 & YOLOv3-416 & 5.183 & 0.017 \\
3 & Faster RCNN (Inception ResNet-v2) & 10.538 & 0.478 \\
4 & YOLOv2-608 & 11.303 & 0.035 \\
5 & Tiny YOLO-416 & 1.018 & 0.011 \\
6 & SSD (Mobilenet v1) & \textbf{0.081} & 0.03 \\
7 & SSD (VGG-300) & 0.361 & \textbf{0.015} \\
8 & SSD (VGG-500) & 0.968 & \textbf{0.026} \\
9 & R-FCN (ResNet-101) & 1.69 & 0.131 \\
10 & Tiny YOLO-608 & 2.144 & \textbf{0.025} \\
11 & SSD (Inception ResNet-v2) & \textbf{0.109} & 0.04 \\
12 & SqueezeDet & \textbf{0.14} & \textbf{0.027} \\
13 & R-FCN & 3.034 & 0.084 \\ \bottomrule
\end{tabular}%
}
\end{table} 

These models benefit relatively simpler deep network with fewer arithmetic operations. This significantly reduced their computational overhead and increased their performance. However, considering the AP result in table 2, it can be inferred that this performance gains, came with an expensive cost of accuracy and precision. Results in GPU mode shows a very similar trend however due to high bandwidth and throughput of GPU, the variance in results are significantly lower. According to Table 3, in GPU mode, SSD (VGG-300), Tiny YOLO-608, and SqueezeDet are among the fastest models in our experiments.
Aside from CPU and GPU latency, we also measured the throughput of continuous inference with repeating image feed. Due to several factors in the experimental setup and model architecture throughput of continuous inference might not be necessarily correlated with the CPU and GPU latency. Figure 3 shows, Tiny YOLO-416 followed by SSD (VGG-300) with over 80 and 60 FPS respectively have the overall highest throughput among the models investigated in this study. On the other hand, Faster RCNN (Inception ResNet-v2) and Faster RCNN (ResNet-101) are slowest in this regard. In order to deploy the deep models in embedded platforms, Caffe or Tensorflow models should be optimized and restructured using Movidius SDK or TensorRT. This enables the CNN model to utilize the target height/width effectively.

\begin{figure}[!h]
  \centering
   {\epsfig{file = 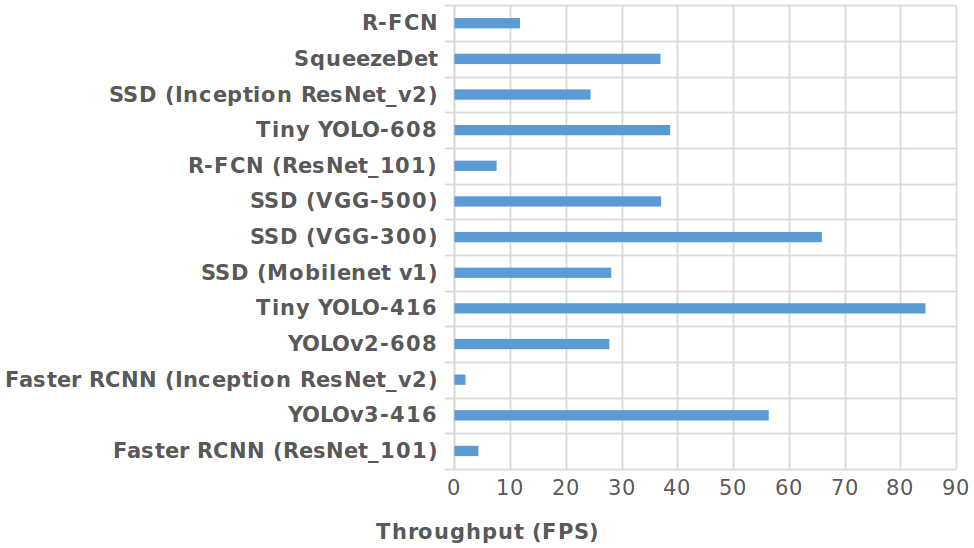, width = 7.5cm}}
  \caption{Throughput of continuous inference across various models}
  \label{fig:example1}
 \end{figure}

However, the supported layers by Movidius SDK or TensorRT are relatively basic and limited and complex
models such as ResNet cannot be truly deployed in these platforms. As an example, leaky rectified linear unit activation function in inception models is not supported by the Jetson platform and cannot be fully replicated. Table 4 summarizes the throughput of continuous inference across various deep models in embedded platforms. It can be observed that the Nvidia Jetson performed significantly better than the Movidius across all different models. Furthermore, TensorRT outperformed Caffe by a relatively large margin. However, in terms of features and functionality, Caffe allows to reproduce more complex networks.

Finding the right deep model for embedded platform is not about accuracy neither performance but is about finding the right tradeoff between accuracy and performance, which satisfies the requirements. Deep models such as Tiny-YOLO can be extremely fast. However, their accuracy is questionable. Figure 4 plots the deep models Average Precision across their throughput. The closer to the top right corner of the plot, the better the overall performance of the model. Figure 4 shows among the various models that we investigated in this research, YOLO v3-416 and SSD (VGG-500) are the best tradeoff between Average precision and throughput.

\begin{table}[]
\caption{Throughput of continuous inference across various models using embedded platform including  Movidius and Jetson}
\label{my-label}
\resizebox{\textwidth/2}{!}{%
\begin{tabular}{@{}lllllll@{}}
\toprule
\multicolumn{1}{c}{\#} & \multicolumn{1}{c}{Model} & \multicolumn{1}{c}{Framework} & \multicolumn{1}{c}{Movidius} & \multicolumn{1}{c}{Jetson} & \multicolumn{1}{c}{} & \multicolumn{1}{c}{} \\ \midrule
\multicolumn{1}{c}{} & \multicolumn{1}{c}{} & \multicolumn{1}{c}{} & \multicolumn{1}{c}{} & \multicolumn{1}{c}{Caffe} & \multicolumn{2}{c}{TensorRT} \\ \cmidrule(l){4-7} 
\multicolumn{1}{c}{} & \multicolumn{1}{c}{} & \multicolumn{1}{c}{} & \multicolumn{1}{c}{\begin{tabular}[c]{@{}c@{}}(FP16)\\ Throughput\end{tabular}} & \multicolumn{1}{c}{\begin{tabular}[c]{@{}c@{}}(FP32)\\ Throughput\end{tabular}} & \multicolumn{1}{c}{\begin{tabular}[c]{@{}c@{}}(FP16)\\ Throughput\end{tabular}} & \multicolumn{1}{c}{\begin{tabular}[c]{@{}c@{}}(FP32)\\ Throughput\end{tabular}} \\ \midrule
1 & AgeNet & Caffe & 18 & 56 & 192 & 127 \\
2 & AlexNet & Caffe & 10 & 37 & 65 & 54 \\
3 & GenderNet & Caffe & 18 & 62 & 198 & 119 \\
4 & GoogleNet & Caffe & 9 & 19 & 120 & 73 \\
5 & SqueezeNet & Caffe & 17 & 37 & 166 & 124 \\
6 & TinyYolo & Caffe & 7 & 19 & -NA- & -NA- \\
7 & Inception v1 & Tensorflow & 10 & -NA- & -NA- & -NA- \\
8 & Inception v2 & Tensorflow & 7 & -NA- & -NA- & -NA- \\
9 & Inception v3 & Tensorflow & 3 & -NA- & -NA- & -NA- \\
10 & Mobilenet & Tensorflow & 19 & -NA- & -NA- & -NA- \\ \bottomrule
\end{tabular}%
}
\end{table}

\begin{figure}[!h]
  \centering
   {\epsfig{file = 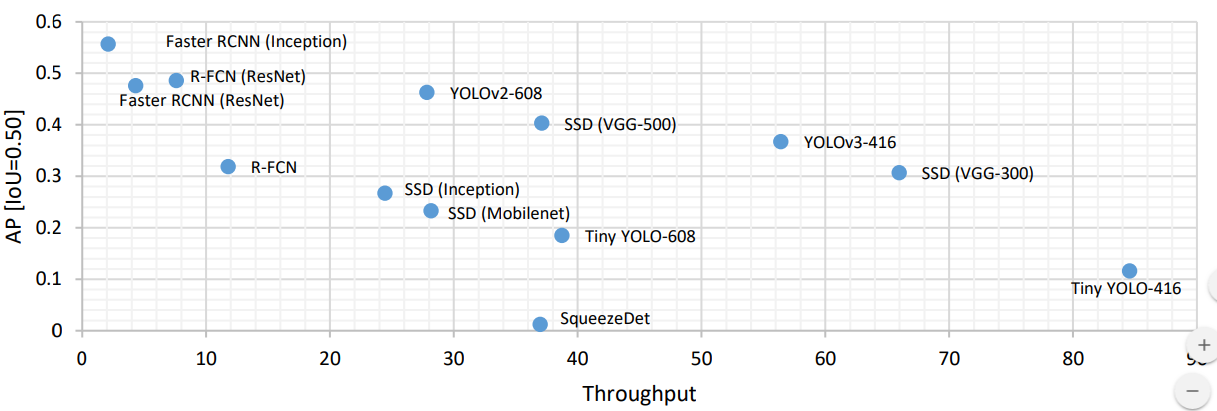, width = 7.5cm}}
  \caption{Average Precision [IoU=0.5] across throughput}
  \label{fig:example1}
 \end{figure}

\section{\uppercase{Conclusion}}
\noindent Person detection is essential step in analysis of the in-store customer behavior and modeling. This study focused on the use of DNN based object detection models for person detection in indoor retail environments using embedded platforms such as the Nvidia Jetson TX2 and the Movidius. Several DNN models including variations of YOLO, SSD, RCNN, R-FCN and SqueezeDet have been analyzed over our proprietary dataset that consists of over 10 thousands images in terms of both time complexity and average precision. Experiments results shows that Tiny YOLO-416 and SSD (VGG-300) are among the fastest models and Faster RCNN (Inception ResNet-v2) and R-FCN (ResNet-101) are the most accurate ones. However, neither of these models nail the tradeoff between speed and accuracy. Further analysis indicates that YOLO v3-416 delivers relatively accurate result in reasonable amount of time, which makes it a desirable model for person detection in embedded platforms. 
\section*{\uppercase{Acknowledgements}}
\noindent 
We thank our colleagues from VCA Technology who provided data and expertise that greatly assisted
the research. This work is co-funded by the EU-H2020 within the MONICA project under grant
agreement number 732350. The Titan X Pascal used for this research was donated by NVIDIA.

\bibliography{ref.bib}
\vfill
\end{document}